# Title: **Automatic and Manual Segmentation of Hippocampus in Epileptic Patients MRI**

Authors: **Mohammad-Parsa Hosseini[1,2], Mohammad-Reza Nazem-Zadeh[2], Dario Pompili[1], Kourosh Jafari-Khouzani[2,3], Kost Elisevich[4,5], Hamid Soltanian-Zadeh[2,6]**

[1] Department of Electrical and Computer Engineering, Rutgers University, New Brunswick, NJ, USA
Parsa@cac.rutgers.edu, pompili@cac.rutgers.edu

[2] Medical Image Analysis Laboratory, Departments of Radiology, Henry Ford Health System, Detroit, MI, USA
Hsoltan1@hfhs.org, mohamadn@rad.hfh.edu

[3] Department of Radiology, Massachusetts General Hospital, Harvard Medical School, Boston, MA, USA
kjafari@nmr.mgh.harvard.edu

[4] Department of Clinical Neuroscience, Spectrum Health System, Grand Rapids, MI, USA
Konstantin.Elisevich@spectrumhealth.org

[5] Division of Neurosurgery, Michigan State University, College of Human Medicine, Grand Rapids, MI, USA

[6] School of Electrical and Computer Engineering, University of Tehran, Tehran, Iran
hszadeh@ut.ac.ir

**Hypothesis:** There is consensus that manual measurement of hippocampal volume is more accurate than the automated techniques because of better visual definition of the hippocampal margins [1]. It remains the standard by which to judge other methodologies [2]. Automated methods, on the other hand, promote operator independence, higher reproducibility and improved clinical applicability [3]. Their application will be judged valid if they provide results comparable to those obtained by manual contouring [4].

**Introduction:** The hippocampus is a seminal structure in the most common surgically-treated form of epilepsy. Accurate segmentation of the hippocampus aids in establishing asymmetry regarding size and signal characteristics in order to disclose the likely site of epileptogenicity. With sufficient refinement, it may ultimately aid in the avoidance of invasive monitoring with its expense and risk for the patient. To this end, a reliable and consistent method for segmentation [5], [6] of the hippocampus from magnetic resonance imaging (MRI) is needed. In this work, we present a systematic and statistical analysis approach for evaluation of automated segmentation methods in order to establish one that reliably approximates the results achieved by manual tracing of the hippocampus.

**Methods:** Hippocampi were manually segmented in 195 mesial temporal lobe epilepsy (mTLE) patients by a previously established protocol using coronal T1-weighted MR images. The ROIs encompassing the hippocampi were outlined in the coronal plane with fine-tuning performed using the sagittal view. The hippocampal position was established using an MRI atlas as reference. Automatic segmentation with FreeSurfer [7] version 5.3.0 was applied which allowed transformation into Talairach coordinates, segmentation of the subcortical white matter and gray matter and removal of nonbrain tissues. Moreover, ABSS [8], a pattern-recognition algorithm, was applied for hippocampal segmentation which shape and signed-distance functions of the manually-established hippocampus were represented on different scales to train Artificial Neural Networks (ANNs). The trained ANNs were applied to extract both right and left hippocampi. The accuracy of different automated segmentation techniques (Freesurfer, ABSS) was assessed by comparison of their results with that of manual segmentation. Intra- and inter-expert variability of segmentation and quality evaluation were considered. Several performance metrics, widely used in the literature, were considered to assess correlation and overlap between automated and manual segmentation results.

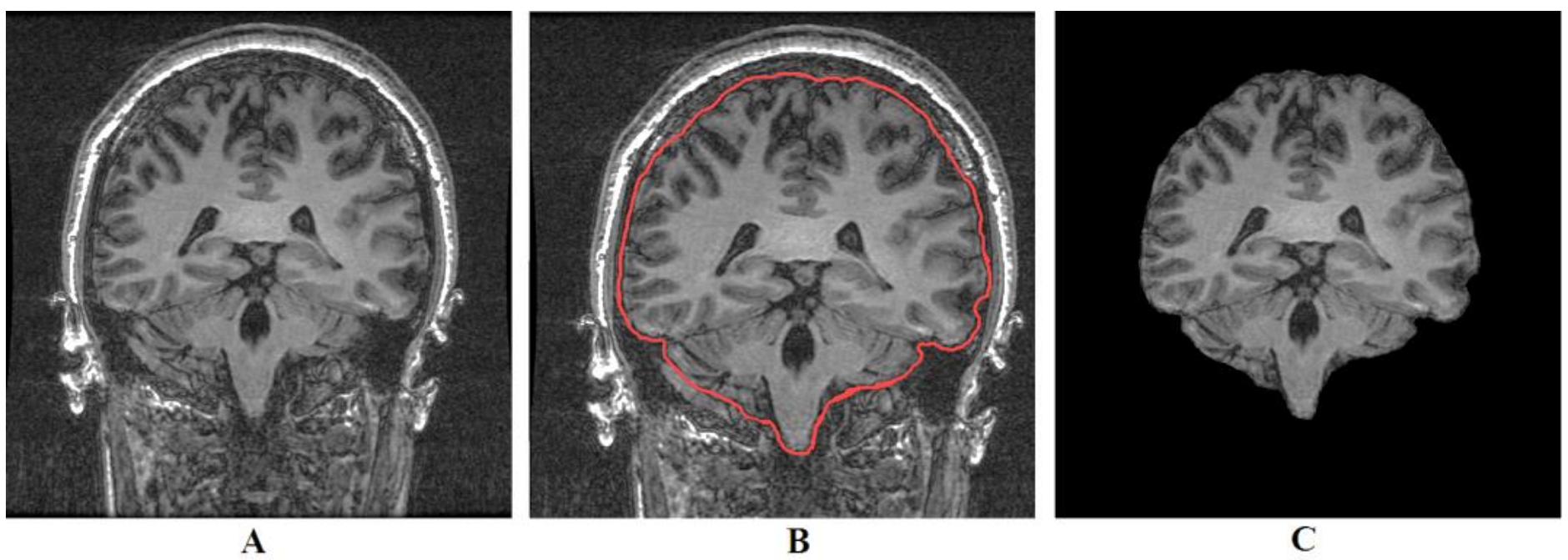

Figure 1: Skull-stripping steps: (A) input images, (B) brain contouring, and (C) removal of nonbrain tissues.

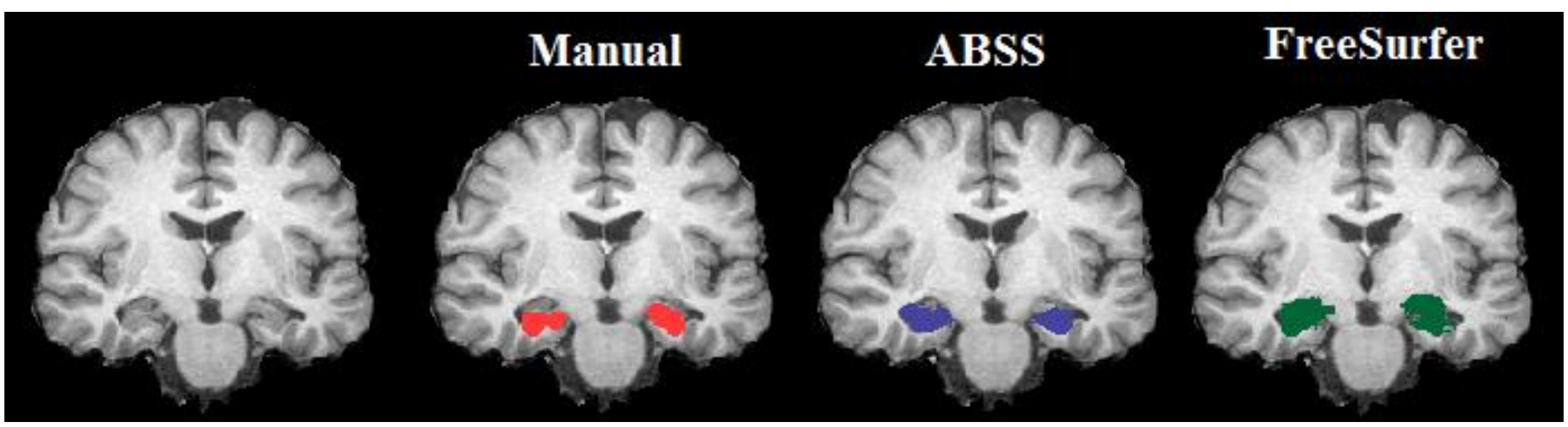

Figure 2: A comparison of manual and automatic hippocampal segmentation methods. The figure shows an intermediate section of an intermediate section of a skull-stripped T1 image of a mTLE patient and the result of manual and automatic segmentations.

**Results:** An archival review of mTLE patients treated between June 1993 and June 2014 at Henry Ford Hospital, Detroit, MI was performed. The patients were evaluated by neurologic examination, video-electroencephalography (EEG), MR and nuclear medicine imaging, and neuropsychological testing. A template database of MR images of 195 mTLE patients (81 males, 114 females; age range of 32-67 years, mean age of 49.16 years) was used in the study. Figure 1 shows the skull-stripping procedure on an intermediate section of a skull-stripped T1 image of a mTLE. Different performance measurements were extracted and calculated for each of the segmentation methods from the T1-weighted images for which Dice, Hausdorff, Precision, and RMS produced values that were felt to fulfill a reasonable basis for analysis. The Dice coefficient for ABSS is 14.10% (p-value$<5\times10^{-33}$) higher compared to FreeSurfer. The ABSS method of segmentation, therefore, is shown to have more overlap with the manual standard than the other methods. The Hausdorff distance for ABSS is 86.73% (p-value$<7\times10^{-19}$) lower compared to FreeSurfer also confirming that the ABSS method better approximates the manual standard. The RMS distance for ABSS is 61.90% (p-value$<6\times10^{-18}$) lower compared to FreeSurfer, demonstrating less variation in values than the other competing methods. Figure 2 shows the result of segmentation methods on an intermediate section of a skull-stripped T1 image of a mTLE patient.

**Discussion and Conclusion:** Automated techniques for hippocampal segmentation have been developed in the research community to reduce time-consuming workload and improve upon reproducibility attributable to the variability encountered in the manual method. A reliable, objective and reproducible technique for automated hippocampal segmentation, in particular, would expedite the confident processing of absolute volumes in clinical cases in order to judge the degree of bihemispheric asymmetry and establish the degree of atrophy over time. Analysis of performance metrics shows that ABSS is a more accurate segmentation method in the case of mTLE.

**Acknowledgment:** A preliminary version of this work appeared in the Proc. of the IEEE Engineering in Medicine and Biology Society (EMBC), Chicago, USA, 2014 [9]. The extended version of this work is published in the International Journal of Medical Physics Research and Practice (Medical Physics), 2016 [10].